\documentclass[conference]{IEEEtran}
\IEEEoverridecommandlockouts

\usepackage{cite}
\usepackage{amsmath,amssymb,amsfonts}
\usepackage{algorithmic}
\usepackage{graphicx}
\usepackage{textcomp}
\usepackage{xcolor}
\usepackage{balance}

\usepackage{etoolbox} 

\renewcommand{\tablename}{TABLE}

\makeatletter
\ifCLASSOPTIONcompsoc
\else
  \patchcmd{\@makecaption}
    {{\normalfont\footnotesize #1}\\{\normalfont\footnotesize\scshape #2}}
    {{\normalfont\footnotesize #1:\ {#2}}} 
    {\typeout{*** Successfully patched \string\@makecaption for 'Table I: caption' format. ***}}
    {\typeout{*** WARNING: Failed to patch \string\@makecaption for 'Table I: caption' format. Check .log. ***}}
\fi
\makeatother

\usepackage{multirow}
\usepackage{booktabs}
\usepackage{xspace}
\usepackage{url}
\newcommand\tokenverse{\texttt{TokenVerse}\xspace}
\newcommand\tokenverseplusplus{\texttt{TokenVerse++}\xspace}
\newcommand\scd{\texttt{[SCD]}\xspace}
\newcommand\epoint{\texttt{[ENDP]}\xspace}
\newcommand\ner{\texttt{[NE]}\xspace}

\newcommand\liden{\texttt{[EN]}\xspace}
\newcommand\lidfr{\texttt{[FR]}\xspace}

\def\BibTeX{{\rm B\kern-.05em{\sc i\kern-.025em b}\kern-.08em
    T\kern-.1667em\lower.7ex\hbox{E}\kern-.125emX}}
\begin{document}

\title{TokenVerse++: Towards Flexible Multitask Learning with Dynamic Task Activation
\thanks{This work was supported by the Idiap~\&~Uniphore collaboration project.
Part of the work was also supported by EU Horizon 2020 project ELOQUENCE (grant number 101070558).
}}



\author{
    \parbox{\linewidth}{%
        Shashi Kumar$^{1,2,\small{\clubsuit}}$, 
        Srikanth Madikeri$^{3}$, Esaú Villatoro-Tello$^{1}$,  
        Sergio Burdisso$^{1}$, Pradeep Rangappa$^{1}$, Andrés Carofilis$^{1}$, Petr Motlicek$^{1,4}$, Karthik Pandia$^{5}$, Shankar Venkatesan$^{5}$, Kadri Hacioğlu$^{5}$ 
        \centering and Andreas Stolcke$^{5}$ \\[2ex]
        \small $^{1}$\textit{Idiap Research Institute, Switzerland};
        \small $^{2}$\textit{EPFL, Switzerland}; 
        \small $^{3}$\textit{University of Zurich, Switzerland};\\ 
        \small $^{4}$\textit{Brno University of Technology, Czech Republic};
        \small $^{5}$\textit{Uniphore, U.S.A. \& India}
        \\\small $^\clubsuit$Corresponding author: \texttt{shashi.kumar@idiap.ch}
    }
}

\maketitle

\begin{abstract}
Token-based multitasking frameworks like TokenVerse require all training utterances to have labels for all tasks, hindering their ability to leverage partially annotated datasets and scale effectively.
We propose TokenVerse++, which introduces learnable vectors in the acoustic embedding space of the XLSR-Transducer ASR model for dynamic task activation. 
This core mechanism enables training with utterances labeled for only a subset of tasks, a key advantage over TokenVerse.
We demonstrate this by successfully integrating a dataset with partial labels, specifically for ASR and an additional task, language identification, improving overall performance.
TokenVerse++ achieves results on par with or exceeding TokenVerse across multiple tasks, establishing it as a more practical multitask alternative without sacrificing ASR performance.
\end{abstract}


\section{Introduction}
\label{sec:intro}
Multitask learning enhances automatic speech recognition (ASR) by enabling multiple tasks in a single inference step, improving efficiency and functionality. Recent token-based approaches \cite{kumar-etal-2024-tokenverse, whisper, ner-e2e-english, wu2024speechcomposer, speechprompt-v2, joint-asr-scd-tag-first, xia2022turn_to_diarize, multitask-asr-scd-icassp} retain conventional ASR architectures while avoiding multi-step pipelines that require separate NLP models for tasks such as named entity recognition \cite{ner-intro} or dialogue modeling \cite{dialogue-call-center-intro, dialogue-modeling-intro}. Unlike large language models \cite{huang2024multilingual, yang2023generative} and speechLLMs \cite{ma2024embarrassingly, das2024speechverse, kumar2025performance} that risk hallucinations, token-based multitask models intersperse task-specific tokens in the hypothesis, preserving the integrity of ASR outputs while augmenting them with additional annotations.

Recently, \tokenverse \cite{kumar-etal-2024-tokenverse} successfully unified four tasks within the XLSR-Transducer \cite{kumar2025xlsr} ASR model. However, its fundamental requirement for exhaustive labeling of all tasks for every training utterance creates a significant bottleneck. This limits its ability to scale by leveraging vast, existing speech corpora that are often only partially annotated, for instance, with ASR transcripts but without named entities or speaker change labels.  This makes the addition of new tasks prohibitively expensive, requiring re-annotation of all data. Although PromptASR \cite{promptASR} partially addresses aspects of task flexibility, it can introduce unnecessary complexity for managing diverse data sources. LLM-inspired methods like prefix-tuning \cite{li-liang-2021-prefix} and prompt-tuning \cite{liu2023gptunderstands} are not directly applicable to ASR models.

Here, we build on the \tokenverse framework and introduce \tokenverseplusplus. It overcomes the task annotation bottleneck through dynamic task activation during training, achieved by learning task-dependent acoustic embeddings within the XLSR-Transducer. It can utilize datasets annotated for only a subset of tasks, ensuring applicability across varied datasets. Experimental results show that \tokenverseplusplus matches or outperforms the original \tokenverse while offering greater flexibility, demonstrating its potential as an efficient solution for token-based multitask ASR.

\section{From TokenVerse to TokenVerse++}
\label{sec:tokenverse++}
\subsection{TokenVerse: Foundations and Limitations}
\label{subsec:tokenverse-recap}
\begin{figure*}[ht]
    \centering
    \includegraphics[width=1.0\linewidth]{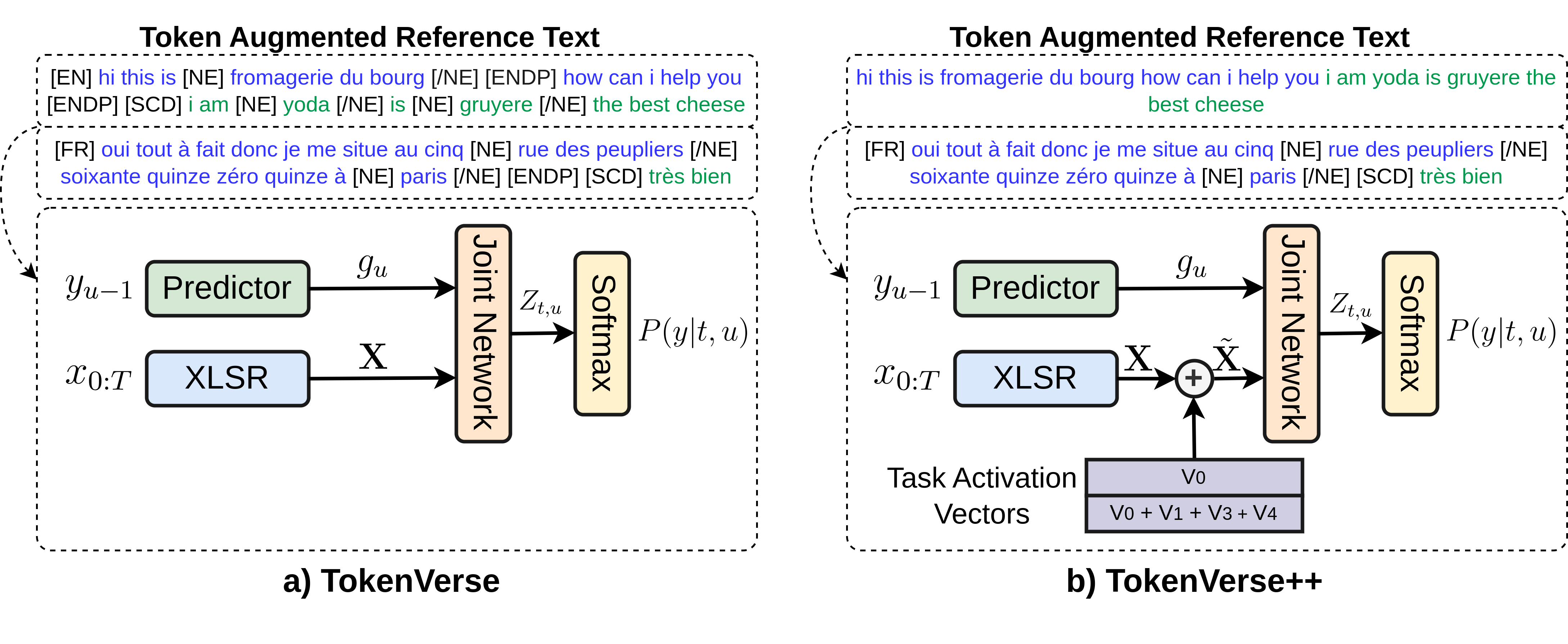}
    \vspace{-1.2cm}
    \caption{Comparison of (a) the original TokenVerse model and (b) the proposed TokenVerse++ framework. TokenVerse (a) requires all training utterances to be fully annotated for all tasks. TokenVerse++ (b) introduces dynamic task activation by adding a learnable task-specific vector \(\mathbf{v}\) (e.g., formed by summing vectors for active tasks like ASR (\(\mathbf{v_0}\)) and auxiliary tasks (\(\mathbf{v_1}: \text{SCD}, \mathbf{v_3}: \text{NER}, \mathbf{v_4}: \text{LID}\))) to the acoustic embeddings from the XLSR encoder. This enables flexible training with partially annotated utterances, allowing different task combinations per input.}
    \label{fig:main-figure}
\vspace{-0.1cm}
\end{figure*}

Token-based multitasking has shown success in \tokenverse \cite{kumar-etal-2024-tokenverse}, even when scaled to four tasks: ASR (the primary task), speaker change detection (SCD), named entity recognition (NER), and endpointing, as illustrated in Figure~\ref{fig:main-figure}(a).
However, training the XLSR-Transducer \cite{kumar2025xlsr} and its transducer loss function \cite{pruned-rnnt-loss,graves2012sequence} as the base ASR model requires labels for all tasks for every training utterance.
This represents a significant limitation, as scaling the number of tasks becomes increasingly difficult, since obtaining gold annotations for multiple tasks for all audio data is challenging.
Additionally, \tokenverse cannot leverage the large number of existing datasets for tasks like ASR, further limiting its applicability.

\subsection{Proposed Approach: TokenVerse++}
\label{subsec:proposed}
The aim of \tokenverseplusplus is to learn a task-aware acoustic embedding space within the \tokenverse model (see Figure~\ref{fig:main-figure}). This is achieved by introducing a mechanism to modulate the original acoustic embeddings using a set of learnable vectors, whose composition based on active tasks is detailed in Section~\ref{subsec:design-choices}.
Let \(\mathbf{X} \in \mathbb{R}^{T \times d}\) denote the acoustic embeddings from the base ASR model, where \(T\) is the number of time steps and \(d\) the embedding dimension. We define a general function \( f: \mathbb{R}^{T \times d} \times \mathbb{R}^{d} \to \mathbb{R}^{T \times d} \) that modifies these embeddings using a dynamically formed, learnable vector \(\mathbf{v} \in \mathbb{R}^{d}\) representing the active task(s):
\begin{equation}
    \tilde{\mathbf{X}} = f(\mathbf{X}, \mathbf{v}).
\end{equation}
In our approach, we hypothesize that a simple element-wise addition is sufficient to effectively condition the embeddings:
\begin{equation}
    \label{eq:vector_addition}
    \tilde{\mathbf{X}} = \mathbf{X} + \mathbf{v}.
\end{equation}
This formulation allows for a straightforward incorporation of task-specific information. We posit that this learned vector \(\mathbf{v}\) acts as a task-specific bias or translation in the acoustic embedding space. By shifting the original embeddings \(\mathbf{X}\), the model can project them into a region of the embedding space that is more conducive to the currently active set of tasks. Thus, the addition of \(\mathbf{v}\) effectively creates a task-conditioned view \(\tilde{\mathbf{X}}\) of the acoustic features, enabling the subsequent layers of the XLSR-Transducer to better differentiate and specialize for the activated tasks.

Specifically, for the XLSR-Transducer ASR model used in \tokenverse, we investigate three configurations for incorporating this task activation into the acoustic embedding space: (i) after the feature encoder, (ii) after the full XLSR \cite{xlsr} encoder, and (iii) a combination of both.


\subsection{Design Considerations in \tokenverseplusplus}
\label{subsec:design-choices}
Assume there are \(K\) tasks apart from a designated always-active, primary task (ASR in our setup).
Auxiliary tasks can be activated independently alongside this primary task. Let \(\mathcal{C}\) represent the set of all possible combinations of auxiliary tasks. This set also includes the scenario where no auxiliary tasks are active, meaning only the primary task is engaged.

We first propose an approach where each unique task combination \( \mathbf{c}_i \in \mathcal{C} \) is associated with its own distinct learnable vector. Here, \(\mathbf{c}_i\) denotes the set of active tasks for a given utterance. To activate a specific subset of tasks \(\mathbf{c}_i\) during training or inference, its corresponding vector is utilized.
For training, a combination \( \mathbf{c}_i \) is selected for each utterance, uniformly sampled from all valid combinations \( \mathcal{C} \), or determined by the available labels for an utterance. 
Observe that the number of combinations ($|\mathcal{C}|$), and consequently the number of learnable vectors, grows exponentially with \(K\), which introduces a large number of trainable parameters as \(K\) becomes larger.

Building on this, we propose a more scalable solution. In this second approach, each individual task \(k\), including the primary task, is associated with a single learnable vector \(\mathbf{v}_k \in \mathbb{R}^d\), where \(d\) is the dimension of the acoustic embedding, rather than having a separate learnable vector for each combination in the set \(\mathcal{C}\). The tasks to be activated are still selected from \(\mathcal{C}\) with equal probability.
When a specific task combination \(\mathbf{c}_i\) needs to be activated, the final activation vector \(\mathbf{v}\) is obtained by summing the vectors corresponding to all tasks present in \(\mathbf{c}_i\), giving
\begin{equation}
    \label{eq:t++-scalable}
    \tilde{\mathbf{X}} = \mathbf{X} + \sum_{k \in \mathbf{c}_i} \mathbf{v}_k
\end{equation}
where each \(\mathbf{v}_k \in \mathbb{R}^d\) is the learnable vector for task \(k\); these vectors are initialized randomly.\footnote{We also tried initializing via approximate determinantal point process (DPP)~\cite{gautier2019two} sampling on a high-dimensional hypersphere, which showed no improvement over random initialization.}
With this method, the number of learnable vectors scales linearly with the total number of tasks. During training, the selection of \(\mathbf{c}_i\) for an utterance typically corresponds to the set of tasks for which labels are available for that particular utterance.

We explore both methods to better understand the trade-off between representational capacity and scalability. The second approach, using a single vector per task, significantly improves scalability.
However, it may reduce task representation granularity due to the many-to-one mapping resulting from the summation.
In contrast, the first approach, utilizing a unique vector per combination, may offer finer-grained task representations but at the cost of a potentially large number of trainable parameters. This trade-off is analyzed in Section~\ref{sec:results}.

{\bf During training}, as illustrated in Figure~\ref{fig:main-figure}(b), a task combination \(\mathbf{c}_i\) is determined for each utterance, typically based on its available annotations. This system allows different utterances within a batch to activate different task sets. These utterances may originate from multiple datasets with varying annotations. This flexibility is a key advantage of \tokenverseplusplus, enabling the model to leverage partially labeled data effectively.

{\bf During inference}, any combination of tasks \(\mathbf{c}_i\) can be activated by selecting or constructing the appropriate task activation vector \(\mathbf{v}\). The model was trained to handle various task combinations at test time, allowing it to work in different application scenarios.

\section{Experimental Setup}
\label{sec:exp-setup}
Our experimental design serves two primary objectives: first, to establish a direct performance comparison between \tokenverseplusplus and the original \tokenverse framework; and second, to show the enhanced flexibility of \tokenverseplusplus through the incorporation of partially annotated datasets and an additional task.

\subsection{Baseline Comparison with TokenVerse}
\label{subsec:exp-baseline}
To ensure a fair and direct comparison with prior work~\cite{kumar-etal-2024-tokenverse}, our initial experiments replicate the setup of the original \tokenverse. This involves focusing on the same four tasks: ASR as the primary task, alongside speaker change detection (SCD), endpointing, and named entity recognition (NER). For this baseline comparison, we use the English portion of the DefinedAI\footnote{\url{https://www.defined.ai/}} dataset, consistent with~\cite{kumar-etal-2024-tokenverse}, which comprises contact center conversations between agents and customers and contains complete annotations for all four tasks. The underlying ASR model remains the XLSR-Transducer~\cite{kumar2025xlsr}, as in \tokenverse. This comparative setup allows for a clear assessment of our proposed dynamic task activation mechanism's impact when all tasks are notionally active during inference, mirroring the \tokenverse evaluation conditions.

\subsection{Leveraging Partially Annotated Datasets}
\label{subsec:exp-expansion-partial}
To evaluate the core capabilities of \tokenverseplusplus, we designed a multilingual, five-task experimental scenario, expanding upon the baseline by introducing language identification (LID) as a fifth task. For LID, language-specific tokens (e.g., \liden, \lidfr) are prepended to the reference transcripts, following the multitask data preparation from~\cite{kumar-etal-2024-tokenverse}.

The training data for this expanded five-task scenario combines fully and partially annotated corpora to test the ability to leverage diverse data sources.
We utilize the multilingual DefinedAI dataset, which provides comprehensive annotations for all five tasks across German (de), Spanish (es), French (fr), and English (en) languages.
Crucially, to evaluate whether \tokenverseplusplus is able to utilize datasets with incomplete task annotations, we integrate a randomly chosen 100-hour subset of the CommonVoice (CV) corpus~\cite{ardila2019_commonvoice_corpus} for each of these four languages.
The size of this specific subset was chosen to provide enough partially labeled data to demonstrate the benefits of our approach, while also ensuring manageable computational demands for training and experimentation across multiple languages and tasks.
Within our five-task framework, the CV dataset provides labels \textit{only} for ASR and LID, thereby lacking annotations for SCD, endpointing, and NER.
As described in Section~\ref{sec:tokenverse++}, \tokenverseplusplus processes such mixed-annotation data by activating only those tasks for which labels are available for any given utterance. In contrast, the original \tokenverse would be unable to utilize these CV utterances due to its strict requirement for complete annotations across all defined tasks.

\begin{table}[t]
\caption{Dataset statistics with token metadata for each subset of the multilingual multitask DefinedAI datasets. Note that the token count for LID equals the number of utterances. In \textbf{Named Entities (NEs) metadata}, \#NE indicates the total count of named entity occurrences, \#uniq represents the number of unique entities among them, and \%Common shows the percentage of named entities shared with the training set for each language.}
\label{tab:dataset-stats}
    \setlength{\tabcolsep}{2.0pt} 
    \begin{tabular}{l cc| ccc| ccc}
    \toprule
    \multicolumn{3}{c|}{\textbf{Datasets metadata}} & \multicolumn{3}{c|}{\textbf{Token metadata [\%]}} & \multicolumn{3}{c}{\textbf{NEs metadata}} \\
    \cmidrule(lr){1-3} \cmidrule(lr){4-6} \cmidrule(lr){7-9}
    subset & \#utt/word  & dur [h] & \scd & \ner & \epoint & \#NE & \#uniq & \%Common \\
    \midrule
    \multicolumn{8}{l}{\textbf{English (en) dataset}} \\
    \cmidrule(lr){2-9}
    train & 10k/359k & 40 & 1.9 & 3.6 & 2.1 & 6.5k & 2350 & -- \\
    test & 1.1k/42k & 4.5 & 1.9 & 3.4 & 2.0 & 727 & 378 & 52.6 \\
    \midrule
    \multicolumn{8}{l}{\textbf{German (de) dataset}} \\
    \cmidrule(lr){2-9}
    train & 15k/487k & 62.8 & 2.6 & 2.3 & 2.7 & 5.6k & 2808 & -- \\
    test & 1.9k/59k & 7.8 & 2.5 & 2.3 & 2.6 & 679 & 454 & 38.3 \\
    \midrule
    \multicolumn{8}{l}{\textbf{Spanish (es) dataset}} \\
    \cmidrule(lr){2-9}
    train & 18k/622k & 73.3 & 2.7 & 2.2 & 2.9 & 7.0k & 2369 & -- \\
    test & 2.1k/73k & 8.7 & 2.8 & 2.2 & 3.0 & 796 & 439  & 52.8 \\
    \midrule
    \multicolumn{8}{l}{\textbf{French (fr) dataset}} \\
    \cmidrule(lr){2-9}
    train & 15k/586k & 63.6 & 2.1 & 1.8 & 2.2 & 5.2k & 1594 & -- \\
    test & 1.8k/68k & 7.4 & 2.1 & 1.7 & 2.2 & 570 & 290 & 49.0 \\
    \bottomrule
    \end{tabular}
\vspace{-0.1cm}
\end{table}
\subsection{Evaluation Metrics}
\label{subsec:exp-metrics}
Performance across the five tasks is measured using established metrics. For ASR, we report word error rate (WER). For SCD and endpointing, we use the text-based F1 score. NER performance is evaluated using the \textit{exact-match} F1 score, consistent with~\cite{kumar-etal-2024-tokenverse}. Finally, LID is assessed by accuracy.

\subsection{Training and Inference Details}
\label{subsec:exp-training}
All \tokenverseplusplus models and baselines are trained using hyperparameters identical to those in the original \tokenverse for consistency.
Specifically, the XLSR-Transducer model~\cite{kumar2025xlsr} is trained with the pruned RNN-T loss~\cite{pruned-rnnt-loss} for $50$ epochs, with an initial learning rate of $1.25\times 10^{-3}$. The best-performing model epoch is selected based on the average WER on the dev sets across all languages. Results are reported on the respective test sets. During inference, decoding is performed using the ``modified\_beam\_search" algorithm with a beam size of 4.

\begin{table*}[t]
    \caption{Performance comparison of \tokenverseplusplus (with all tasks activated at inference) and \tokenverse on the DefinedAI English test set. \tokenverseplusplus results are reported when task activation is applied at different embedding spaces within the XLSR encoder (see Section~\ref{subsec:proposed}). NER F1 scores are calculated using the \textit{exact-match} approach from \cite{kumar-etal-2024-tokenverse}. The best \tokenverseplusplus configuration improves ASR performance, maintains comparable performance in NER and endpointing, but shows a decline in SCD.}
    \label{table:tokenverse-main}
    \centering
    \small
    \begin{tabular}{l cccc}
        \toprule
        \textbf{Model} & \textbf{ASR (WER ↓)} & \textbf{SCD (F1 ↑)} & \textbf{Endpoint (F1 ↑)} & \textbf{NER (F1 ↑)} \\
        \midrule
        \tokenverse~\cite{kumar-etal-2024-tokenverse} & 14.7 & \textbf{90.3} & 89.9 & 57.6 \\
        \midrule
        \multicolumn{5}{l}{\textbf{\tokenverseplusplus (Each Task Combination Has a Learnable Vector)}} \\
        \quad After Feature Encoder & \textbf{14.1} & 88.2 & 89.7 & \textbf{57.7} \\
        \quad After Full XLSR Encoder & 14.7 & 86.3 & 87.6 & 55.4 \\
        \quad After Both & 14.3 & 88.3 & 89.8 & 57.4 \\
        \midrule
        \multicolumn{5}{l}{\textbf{\tokenverseplusplus (Each Task Has a Learnable Vector)}} \\
        \quad After Feature Encoder & 14.5 & 89.4 & 89.4 & 55.4 \\
        \quad After Full XLSR Encoder & 14.4 & 89.3 & \textbf{90.0} & 53.1 \\
        \quad After Both & 14.7 & 88.7 & 89.4 & 56.0 \\
        \bottomrule
    \end{tabular}
    \vspace{5pt}
\end{table*}
\subsection{Dataset Details}
\label{sec:appendix-dataset-details}
Table~\ref{tab:dataset-stats} provides detailed statistics for the multilingual multitask DefinedAI datasets, covering English, German, Spanish, and French, along with token metadata for each subset.
The token count for LID matches the number of utterances, as LID is performed at the utterance level.
Notably, the percentage of unique named entities in test sets unseen during training ranges from 47.2\% to 61.7\%, making the datasets especially valuable for evaluating the generalization capabilities of \tokenverseplusplus. The diversity in label availability, along with the presence of unseen named entities across tasks and languages, offers a comprehensive and robust testbed for multitask learning in a multilingual context.

\section{Results and Discussions}
\label{sec:results}
We first evaluate \tokenverseplusplus against the \tokenverse baseline using fully annotated data. Subsequently, we analyze its core capability of leveraging partially annotated datasets in a multilingual, multitask setting. Finally, we discuss the behavior of the dynamic task activation mechanism during inference and the overall implications of our approach.

\subsection{Baseline Comparison with \tokenverse}
\label{subsec:baseline_comparison}
Our initial investigation assesses \tokenverseplusplus under conditions identical to those of the original \tokenverse evaluation, employing the English DefinedAI dataset with annotations for ASR, SCD, endpointing, and NER. To ensure a direct comparison, all tasks were activated during inference, mirroring \tokenverse's standard inference mode.
As noted before, during training of \tokenverseplusplus, random combinations of tasks will be activated.
Results are presented in Table~\ref{table:tokenverse-main}.

The \textbf{per-combination task vector strategy} (Section~\ref{subsec:design-choices}), when task activation is introduced after the feature encoder, shows an advantage. This configuration yields 4\% relative improvement in ASR WER over \tokenverse while maintaining similar performance in NER and endpointing. A trade-off is observed for SCD performance. Conversely, intervening \textit{after the full XLSR encoder} with this strategy is detrimental, leading to a general decline in performance across all tasks. This suggests that modifying embeddings at this later stage, with fewer subsequent learnable parameters in the encoder, offers insufficient representational power for effective multitask adaptation. Activating tasks at \textit{both} the feature encoder and full XLSR encoder simultaneously does not yield further benefits over the feature encoder-level intervention alone.

The \textbf{more scalable per-task vector summation approach}, where tasks each have independent learnable vectors that are summed to form the final task activation vector (Eq.~\ref{eq:t++-scalable}), the results closely match those of \tokenverse across most tasks. However, a slight degradation is generally observed compared to the per-combination vector strategy. This may be due to the summation operator, which is a many-to-one mapping that may introduce ambiguity for task activation. Notably, this approach shows some improvement in ASR, SCD, and endpointing when tasks are activated after the full XLSR encoder.

These findings have architectural implications. The outcomes achieved with feature-encoder-level intervention, especially for the per-combination strategy, underscore the benefit of early modulation of acoustic embeddings.
This allows the deep XLSR encoder to process task-conditioned inputs, thereby facilitating more effective hierarchical feature learning for diverse tasks. A trade-off emerges: the per-combination strategy achieves the best observed performance at the expense of an exponential increase in task-specific parameters ($2^{K}$ vectors), whereas the per-task summation strategy offers linear scalability ($K$ vectors) with a slight compromise in performance for certain configurations. Given that \tokenverse itself established a strong benchmark by outperforming pipeline methodologies~\cite{kumar-etal-2024-tokenverse}, these results establish \tokenverseplusplus as an effective and flexible successor.

\begin{table*}[t]
    \caption{Performance comparison on multilingual DefinedAI eval sets for \tokenverse, and the best-performing \tokenverseplusplus (with all tasks activated at inference) across four languages. The comparison includes \tokenverseplusplus evaluated with only the multilingual DefinedAI dataset and after the addition of CommonVoice (CV) in training, which provides labels solely for ASR and LID tasks. Notably, \tokenverse would not allow the inclusion of utterances (from the CV dataset in this case) in training that lack labels for all tasks. In contrast, \tokenverseplusplus enables such inclusion, improving performance on such tasks.}
    \label{table:multilingual-comparison}
    \centering
    \renewcommand{\arraystretch}{1.2}
    \setlength{\tabcolsep}{6.6pt}
    \small 
    \begin{tabular}{l cccccc}
        \toprule
        \textbf{Model} & \textbf{Lang} & \textbf{ASR (WER ↓)} & \textbf{SCD (F1 ↑)} & \textbf{Endpoint (F1 ↑)} & \textbf{NER (F1 ↑)} & \textbf{LID (acc ↑)} \\
        \midrule
        \multirow{4}{*}{\tokenverse} & en & 15.2 & 90 & 89.7 & 53.7 & 99.8 \\
        & de & 21.3 & 80.8 & 80.1 & 25.5 & \textbf{99.8} \\
        & es & 24.4 & 66.6 & 66.4 & 42.4 & \textbf{100.0} \\
        & fr & 20.8 & 70.5 & 70.2 & 45.3 & \textbf{99.9} \\
        \midrule
        \multirow{4}{*}{\tokenverseplusplus} & en & 13.9 & 91.1 & \textbf{91.2} & 54.1 & 99.8 \\
        & de & 19.9 & \textbf{82.6} & \textbf{82.1} & \textbf{25.8} & 99.5 \\
        & es & 22.7 & \textbf{66.9} & \textbf{67.5} & \textbf{42.9} & 99.6 \\
        & fr & 18.8 & 73.6 & 73.7 & 47.8 & \textbf{99.9} \\
        \midrule
        \multirow{4}{*}{\tokenverseplusplus~(with CV)} & en & \textbf{13.3} & \textbf{91.3} & 90.4 & \textbf{56.0} & \textbf{100.0} \\
        & de & \textbf{18.8} & 80.9 & 80.4 & 25.1 & 99.4 \\
        & es & \textbf{22.2} & 62.7 & 62.1 & 39.3 & 99.5 \\
        & fr & \textbf{18.1} & \textbf{74.0} & \textbf{74.1} & \textbf{48.6} & \textbf{99.9} \\
        \bottomrule
    \end{tabular}
\end{table*}
\subsection{Leveraging Partially Annotated Datasets}
\label{subsec:results_partial_data_short}
A key aspect of \tokenverseplusplus is its ability to integrate datasets with incomplete task annotations. This capability was explored through a multilingual, five-task scenario (ASR, SCD, endpointing, NER, and LID), as detailed in Section~\ref{subsec:exp-expansion-partial}. The LID task serves here as a practical test case of how \tokenverseplusplus accommodates tasks for which label availability varies across datasets. We employed the \tokenverseplusplus configuration using per-combination vectors with activation after the feature encoder. Results are presented in Table~\ref{table:multilingual-comparison}.

\paragraph{Performance with Fully Annotated Multilingual Data (DefinedAI-only)}
Initially, training was conducted solely on the multilingual DefinedAI dataset, which provides annotations for all five tasks. In this setting, \tokenverseplusplus outperforms \tokenverse across the majority of tasks and languages. For instance, ASR WER sees improvements across all evaluated languages, as do metrics for SCD and endpointing. NER performance also shows positive gains. LID accuracy is high for both models. This establishes a five-task baseline before introducing partially labeled data, confirming that \tokenverseplusplus can handle multiple tasks robustly.

\paragraph{Enhancements via Partially Labeled Dataset Integration (DefinedAI + CommonVoice)}
Next, we utilize the CommonVoice (CV) dataset during training in addition to the DefinedAI data. 
The CV data contributes labels \textit{solely} for ASR and LID, thereby representing a corpus of partially annotated speech. As hypothesized, this integration yields further improvements, particularly for ASR. ASR error rates are consistently reduced across all languages, with significant gains observed for German. LID accuracy remains high.
For tasks where CV offered no annotations (SCD, endpointing, NER), performance is largely maintained or shows marginal improvements in several language-task pairs. While isolated, minor degradations in these auxiliary tasks are observed for specific languages upon adding the CV data.
The improvements in ASR performance affirm the utility of leveraging large, partially-annotated ASR corpora. This capability, to enhance performance on tasks by incorporating readily available, albeit incompletely labeled, datasets, is a fundamental advantage of \tokenverseplusplus, unachievable within the constraints of the original \tokenverse framework. The inclusion of LID and its successful training with partial support from CV further underscores this adaptability.

\begin{table*}[t]
    \caption{Performance of \tokenverseplusplus for two configurations: (i) each task combination has a learnable vector, and (ii) each task has a learnable vector. Results are reported for the best performing activation positions from Table~\ref{table:tokenverse-main}, with different subsets of tasks activated at inference while keeping the primary task (ASR) always active. The \textbf{Inactive Tasks Token (ITT)} column indicates the count of tokens predicted for tasks not activated, showcasing the model's effectiveness in minimizing irrelevant predictions.}
    \label{table:appendix-diff-task-activation}
    \centering
    \small
    \setlength{\tabcolsep}{6pt} 
    \renewcommand{\arraystretch}{1.0}
    \begin{tabular}{l cccc c}
        \toprule
        \textbf{Activated Tasks} & \textbf{ASR (WER ↓)} & \textbf{SCD (F1 ↑)} & \textbf{Endpoint (F1 ↑)} & \textbf{NER (F1 ↑)} & \textbf{ITT (↓)} \\
        \midrule
        \multicolumn{6}{l}{\textbf{Each Task Combination Has a Learnable Vector (Activation After Feature Encoder)}} \\
        \midrule
        \textit{Single Task} \\
        \quad ASR & 14.3 & -- & -- & -- & 0 \\
        \quad ASR + SCD & 14.2 & \textbf{89.5} & -- & -- & 0 \\
        \quad ASR + Endpoint & 14.2 & -- & \textbf{90.1} & -- & 0 \\
        \quad ASR + NER & 14.1 & -- & -- & 57.5 & 0 \\
        \midrule
        \textit{Two Tasks} \\
        \quad ASR + SCD + Endpoint & 14.2 & 87.7 & 89.4 & -- & 0 \\
        \quad ASR + SCD + NER & 14.2 & \textbf{89.5} & -- & 57.7 & {\color{red}1} \\
        \quad ASR + Endpoint + NER & \textbf{13.9} & -- & 90.0 & \textbf{58.3} & {\color{red}1} \\
        \midrule
        \textit{All Tasks} \\
        \quad ASR + SCD + Endpoint + NER & 14.1 & 88.2 & 89.7 & 57.7 & -- \\
        \midrule
        \multicolumn{6}{l}{\textbf{Each Task Has a Learnable Vector (Activation After Feature Encoder)}} \\
        \midrule
        \textit{Single Task} \\
        \quad ASR & 14.7 & -- & -- & -- & 0 \\
        \quad ASR + SCD & 14.7 & 87.8 & -- & -- & 0 \\
        \quad ASR + Endpoint & 14.7 & -- & 88.6 & -- & 0 \\
        \quad ASR + NER & \textbf{14.5} & -- & -- & 55.3 & 0 \\
        \midrule
        \textit{Two Tasks} \\
        \quad ASR + SCD + Endpoint & 14.6 & 88.9 & 88.9 & -- & 0 \\
        \quad ASR + SCD + NER & \textbf{14.5} & 88.1 & -- & \textbf{55.5} & {\color{red}1} \\
        \quad ASR + Endpoint + NER & \textbf{14.5} & -- & 88.6 & 55.1 & 0 \\
        \midrule
        \textit{All Tasks} \\
        \quad ASR + SCD + Endpoint + NER & \textbf{14.5} & \textbf{89.4} & \textbf{89.4} & 55.4 & -- \\
        \bottomrule
    \end{tabular}
    \vspace{-0.1cm}
\end{table*}

\subsection{Analyzing Dynamic Task Activation Performance}
\label{sec:appendix-moto-results}
The adaptability of the dynamic task activation mechanism is further explored by evaluating \tokenverseplusplus with varying subsets of active tasks during inference, on the English DefinedAI dataset.
These results, detailed in Table~\ref{table:appendix-diff-task-activation}, present inter-task dynamics and differential impacts of the per-combination versus per-task summed activation vector strategies.

\paragraph{Per-Combination Task Vector Strategy}
This strategy, using unique vectors per task combination, shows that ASR performance is sensitive to co-activated tasks. ASR WER generally improves when an auxiliary task (SCD, endpointing, or particularly NER) is co-activated, compared to ASR-only activation. The combination of \texttt{ASR+Endpointing+NER} yields the best ASR and NER performance in these experiments, suggesting strong synergy where the specialized vector benefits all three. Conversely, auxiliary tasks like SCD and endpointing achieve their individual peak F1-scores in smaller, more focused combinations (e.g., ASR with only SCD or only endpointing), indicating that their optimal conditioning might be achieved without the influence of all other tasks.

\paragraph{Per-Task Vector Summation Strategy}
Summing individual task vectors results in more stable ASR performance across different task combinations, often near its level when co-activated with NER or all tasks. This suggests a more generalized ASR conditioning. Notably, auxiliary tasks like SCD and endpointing tend to reach their highest F1-scores when \textit{all} tasks are active, contrasting with the per-combination strategy. This implies that the cumulative effect of summing multiple task vectors can be beneficial for these specific auxiliary tasks in a broadly active setting.

\paragraph{Comparative Insights}
The \textbf{per-combination strategy} excels at optimizing for specific task subsets, potentially capturing specialized synergies (e.g., \texttt{ASR+Endpointing+NER}). Its tailored vectors can lead to peak performance for these targeted combinations. The \textbf{per-task summation strategy} offers more distributed task influence. While it may not achieve the same peak ASR synergy, its additive nature can be advantageous for certain auxiliary tasks when many tasks are active, providing a robust, general conditioning.

In essence, the choice of strategy for the task activation vector dictates how inter-task dependencies are manifested and exploited. The per-combination approach allows for fine-tuned optimization for specific task groupings, while the per-task summation provides a more general, scalable mechanism with different optimal activation conditions for auxiliary tasks.


\subsection{Impact}
\label{subsec:impact_implications}
\tokenverseplusplus demonstrates that ASR models can be trained effectively to perform multiple speech labeling tasks, utilizing both existing ASR datasets and additional datasets annotated for other tasks. This approach improves ASR performance compared to standalone ASR models and \tokenverse, offering a more practical solution. Since \tokenverseplusplus can predict ASR-type hypotheses while incorporating additional tasks in a single inference step, it is an attractive replacement for ASR-only models, particularly within the transducer architecture.

\section{Conclusion}
\label{sec:conclusion}
We have introduced \tokenverseplusplus, a token-based multitask framework that overcomes the data inflexibility of prior models. By incorporating learnable vectors for dynamic task activation within the XLSR-Transducer, \tokenverseplusplus successfully leverages existing ASR datasets alongside additional task-specific datasets, even those with only partial annotations. Our experiments demonstrated that \tokenverseplusplus achieves results on par with, or exceeding, the original \tokenverse across multiple tasks, notably improving ASR performance. Crucially, by integrating partially labeled data, specifically, the CommonVoice corpus for ASR and language identification as an additional task, we further enhanced ASR performance, showcasing the system's ability to effectively utilize diverse data sources. The flexible task activation scheme in training underscores the potential of our approach for unified modeling across diverse speech and NLP tasks, offering a more practical and adaptable solution for multitask learning.




\clearpage
\balance
\bibliographystyle{IEEEtran}
\bibliography{custom}


\end{document}